\documentclass[10pt, journal, lettersize]{IEEEtran}
\IEEEoverridecommandlockouts
\usepackage{xspace,amsmath,amssymb,amsfonts,epsfig,subfigure,syntonly}
\usepackage[skip=3pt]{caption}
\usepackage{lipsum}
\usepackage{cite,bm,color,url,textcomp,empheq,boxedminipage}
\usepackage{algorithmicx}
\usepackage{indentfirst}
\usepackage{epstopdf,makecell}
\usepackage{empheq}
\usepackage{multicol}
\usepackage{pifont}
\usepackage{cuted}  
\usepackage{booktabs}
\usepackage{stfloats}
\usepackage[noend]{algpseudocode}
\usepackage{tcolorbox}
\usepackage{mdframed}
\usepackage[linesnumbered,ruled]{algorithm2e}
\usepackage{graphicx,graphics}  
\usepackage{multirow,multicol}
\usepackage{psfrag}    
\usepackage{stfloats}
\usepackage{url}
\usepackage{lipsum}

\newcommand\blfootnote[1]{%
  \begingroup
  \renewcommand\thefootnote{}\footnote{#1}%
  \addtocounter{footnote}{-1}%
  \endgroup
}

\long\def\symbolfootnote[#1]#2{\begingroup
\def\thefootnote{\fnsymbol{footnote}}
\footnote[#1]{#2}\endgroup}
\psfull

\allowdisplaybreaks[4]
\begin{document}

\title{ComAgent: Multi-LLM based Agentic AI Empowered Intelligent Wireless Networks}

\author{Haoyun Li, Ming Xiao, \emph{Senior Member, IEEE}, Kezhi Wang, \emph{Senior Member, IEEE}, Robert Schober, \emph{Fellow, IEEE}, Dong In Kim, \emph{Life Fellow, IEEE}, and Yong Liang Guan, \emph{Senior Member, IEEE} }

\maketitle

\begin{abstract}
Emerging wireless networks such as the sixth-generation (6G) mobile systems are expected to operate over highly heterogeneous infrastructures with rapidly evolving service demands, thereby increasing the dependence on large-scale, constraint-coupled, cross-layer optimization for network design and operation. However, the trend leads to a fundamental bottleneck: Converting high-level intents into mathematically consistent formulations, implementable algorithms, and reproducible simulations remains predominantly human-driven, time-intensive, and error-prone. While large language models (LLMs) provide a natural-language interface for intent interpretation and rapid prototyping, monolithic LLM-based pipelines are often limited by insufficient domain grounding, weak constraint awareness, and a lack of execution-based verification and self-correction. These limitations motivate a shift towards agentic AI, where problem-solving is realized through iterative decomposition, explicit planning, tool-integrated execution, and reflection driven by feedback. In this paper, we present ComAgent, a multi-LLM based agentic AI framework that coordinates specialized agents for literature searching, planning, coding, and solution scoring within a closed-loop Perception–Planning–Action–Reflection cycle, turning user intents into solver-ready formulations and reproducible simulation pipelines while continuously self-correcting logical and feasibility errors. We demonstrate the efficacy of ComAgent through two distinct evaluations. In a non-trivial beamforming optimization case study, ComAgent autonomously perceives the problem, designs an algorithm, and generates solutions that achieve a performance comparable to that of expert-designed baselines. Furthermore, on a diverse set of generic wireless tasks, ComAgent outperforms monolithic LLMs. The numerical results demonstrate the potential of the proposed agentic AI framework for various emerging wireless networks.
\end{abstract}
\begin{IEEEkeywords}
Large language models, agentic AI systems, wireless networks, and optimization.
\end{IEEEkeywords}
\blfootnote{H. Li and M. Xiao are with the Division of Information Science and Engineering, KTH Royal Institute of Technology, Stockholm 10044, Sweden (e-mail: \{haoyunl, mingx\}@kth.se). K. Wang is with the Department of Computer Science, Brunel University of London, Uxbridge, Middlesex, UB8 3PH (e-mail: kezhi.wang@brunel.ac.uk). 
Robert Schober is with the Institute for Digital Communications, Friedrich-Alexander-University Erlangen-Nurnberg (FAU), 91054 Erlangen, Germany (e-mail: robert.schober@fau.de). Dong In Kim is with the Department
of Electrical and Computer Engineering, Sungkyunkwan University, Suwon
16419, South Korea (e-mail: dongin@skku.edu). Yong Liang Guan is with Continental-NTU Corporate Lab, Nanyang Technological University, Singapore 639798 (e-mail: EYLGuan@ntu.edu.sg).}

\section{Introduction}
Emerging wireless networks such as the sixth-generation (6G) are envisioned as highly heterogeneous and integrated infrastructures. Key enablers include space-air-ground integrated networks (SAGIN), ultra-massive multi-input-multi-output (MIMO) base stations (BS), reconfigurable intelligent surfaces (RIS), simultaneous wireless information and power transfer (SWIPT), integrated sensing and communication (ISAC), and edge intelligence \cite{survey0}. Furthermore, service demands are shifting from best-effort connectivity to intent-driven applications such as immersive extended reality (XR), autonomous systems, and large-scale Internet of Things (IoT), which impose stringent and often conflicting requirements on data rate, latency, reliability, positioning accuracy, and energy efficiency. The strong coupling across layers and domains poses unprecedented design challenges for resource allocation, topology control, sensing–communication co-design, and cross-layer management.

Meanwhile, the rapid expansion in scale and diversity exposes the fundamental limitations of conventional design workflows. Traditional optimization methods require manually designed models and problem-specific algorithms that exploit specific structures for problem decomposition and non-convexity transformation. However, the complexity of real-world environments and applications results in challenging optimization problems that are high-dimensional, strongly non-convex, and even multi-objective, involving mixed discrete–continuous decisions and complex coupling constraints. Each new scenario typically requires significant expert knowledge to redesign mathematical formulations and tailored algorithms. This workflow substantially restricts the scalability and flexibility of dynamically changing wireless network environments. Although data-driven deep learning (DL) and reinforcement learning (RL) methods can alleviate modeling burdens to certain degrees, they remain restricted by the growing complexity and volume of the required training data, high computational demands, and overfitting risks. These methods require scenario-specific datasets and expert knowledge of networks to formulate optimization tasks, making them time- and resource-consuming to adapt to new tasks and network configurations \cite{MOPsurvey}. Consequently, the main bottleneck is the growing reliance on human domain expertise and the specialized custom algorithm design required to develop end-to-end solutions.

On the other hand, large language models (LLMs) present a paradigm shift for network design, transitioning from rigid mathematical optimization to flexible, semantic-aware control. Leveraging superior generalization, reasoning, and in-context learning capabilities \cite{Chen}, LLMs operate directly on high-level textual specifications, bypassing the need for explicit mathematical modeling, which is common in traditional methods. Unlike specialized algorithms that require retraining, LLMs exhibit zero- or few-shot adaptation to new tasks via in-context examples. Thus, LLMs can unify heterogeneous network functions, such as configuration generation, optimization, and troubleshooting, within a single instruction-driven interface \cite{zhou, wang}. This facilitates an end-to-end network management framework in which natural language directives autonomously drive the synthesis of network configurations and control algorithms, significantly streamlining complex network orchestration. However, integrating monolithic LLMs directly into wireless optimization loops presents fundamental challenges. LLMs are not native numerical optimizers, lack gradient information, struggle with the high-precision, complex-valued arithmetic essential for physical-layer optimization, and often provide no convergence guarantees for typical non-convex problems. Furthermore, a single generic model cannot access dynamic operator-specific knowledge bases or evolving wireless network standards, leading to hallucinations or constraint violations. This monolithic approach is also untenable for real-time edge applications, as specialization across diverse tasks (e.g., retrieval, decomposition, and verification) incurs prohibitive latency and resource overheads. 

These observations motivate a paradigm shift from chat-style LLM assistance to agentic AI, where LLMs are embedded into an autonomous, tool-augmented, feedback-driven decision loop \cite{jiang}. In an agentic AI system, intelligence is not only reflected by text generation, but also by iterative task decomposition, explicit planning, tool invocation (e.g., solvers, simulators, retrieval engines), and reflection based on execution feedback. The agentic loop is particularly well aligned with wireless optimization workflows since action naturally corresponds to running code and invoking optimization modules, while reflection corresponds to diagnosing infeasibility, correcting modeling errors, and revising solution strategies under physical constraints and numerical feedback.

\begin{figure*}[t] 
\centering
 \makebox[\textwidth]{\includegraphics[width=.85\paperwidth]{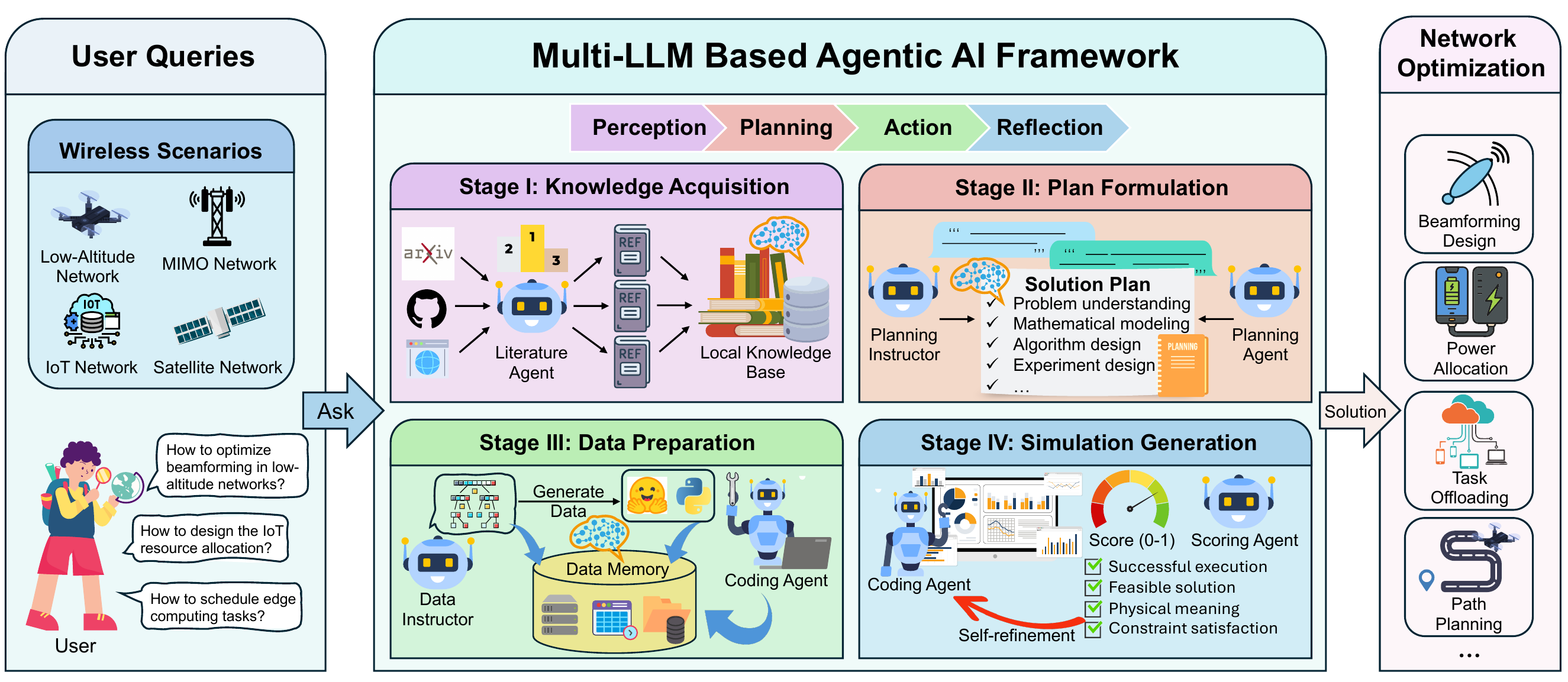}}
\caption{Framework of ComAgent: A multi-LLM based agentic AI system.}
\label{framework}
\end{figure*}

Motivated by this perspective, we present ComAgent, a multi-LLM agentic AI framework that targets end-to-end wireless optimization as an autonomous workflow rather than a standalone model query, as shown in Fig. \ref{framework}. The proposed framework coordinates specialized agents, including Literature, Planning, Coding, and Scoring Agents, within a recursive cognitive cycle to transform high-level intents into rigorous mathematical formulations and executable simulation code. By anchoring decisions in retrieved domain knowledge, enforcing structured task decomposition, and leveraging scoring-based feedback for iterative self-correction, ComAgent aims to ensure constraint satisfaction, physical feasibility, and implementation fidelity throughout the optimization pipeline. We demonstrate the effectiveness of the proposed framework through representative evaluations. In a non-trivial MIMO SWIPT beamforming case study, ComAgent autonomously perceives the task, designs a solution strategy, and generates executable implementations that achieve performance comparable to expert-designed baselines. Furthermore, on a generic test set of diverse wireless tasks, the proposed agentic framework achieves a substantially higher solution success rate than a single LLM, indicating improved robustness and generalization under heterogeneous problem settings. These results suggest that the key value of LLMs for future wireless networks may lie in enabling autonomous orchestration that bridges intent understanding, mathematical rigor, and code-centric validation in a unified closed-loop system.
\section{The Landscape of LLM-Driven Optimization: Progress and Gaps}

\subsection{Recent advances in LLM-enabled wireless optimization}

The integration of LLMs has emerged as a completely new paradigm for intelligent network management, enabling a shift from rigid data-driven DL pipelines to flexible, semantic-aware control. Nevertheless, applying monolithic LLMs directly to high-precision and non-convex wireless optimization problems remains fundamentally challenging. In particular, LLMs are not native numerical optimizers and lack internal mechanisms to guarantee convergence or to satisfy complex, domain-specific constraints. This limits their reliability for engineering-grade closed-loop optimization. This core limitation has inspired two primary approaches to harness the capabilities of LLMs for complex wireless optimization.

The first category involves hybrid frameworks that embed an LLM as a heuristic component within a traditional, human-designed algorithmic pipeline. For instance, the authors of \cite{Li1} proposed the use of an LLM as a black-box solver for the intractable combinatorial optimization of user association while relying on conventional convex optimization methods, including alternating direction method of multipliers (ADMM), majorization minimization (MM), and fractional programming (FP) for numerical beamforming design. Similarly, the authors in \cite{Li2} integrated an LLM as a generative search operator into a multi-objective evolutionary algorithm tasked with producing novel candidate solutions. The key progress of these methods lies in augmenting classical solvers with LLM reasoning to improve adaptability and reduce manual trial-and-error.

Meanwhile, multi-agent architectures have been explored to decompose complex network management tasks by domain, mimicking the division of labor in an operational team. For instance, the authors in \cite{Qayyum} introduced LaMA-SON, a system with specialized agents for discrete network functions, such as traffic, quality of service (QoS), and security. The authors in \cite{cao} designed a similar multi-agent system for situational awareness in zero-trust SAGINs, with agents focused on threat assessment and decision-making. Other works emphasize the cognitive architecture itself, such as \cite{tong}, which proposed a general framework inspired by human cognitive steps (perception, memory, and planning), and \cite{liu}, which leveraged retrieval-augmented generation (RAG) to improve inter-agent communication efficiency. These studies represent important progress toward semantically driven network operations, where agents monitor, classify, and execute operational logic in a scalable manner.

\subsection{Open challenges toward end-to-end agentic optimization copilots}

Despite the above progress, a critical limitation of the proposed hybrid schemes is that LLMs typically function as subordinate modules constrained within rigid, predetermined algorithmic pipelines. As a result, they cannot autonomously reorganize the workflow, redesign the solution strategy, or validate feasibility when the problem context changes. Meanwhile, existing multi-agent systems are primarily designed for high-level network management automation rather than mathematically grounded optimization. Although they enable semantic monitoring and decision support, they are generally not designed to translate intent into rigorous formulations and executable solvers under physical constraints.

A notable attempt in \cite{zhang2} has begun to narrow this gap by applying a generative agent for problem formulation to specify the mathematical objective, while relying on a separate deep RL solver to obtain solutions. While promising, this design still exposes a structural disconnect between formulation and solution generation, and it lacks a unified mechanism for execution-grounded verification. In practical wireless optimization, the decisive bottleneck often lies not only in determining the objective function, but also in producing correct and reproducible implementations that satisfy constraints and remain physically meaningful after numerical execution. These observations highlight the absence of a single integrated framework that unifies knowledge grounding, problem formulation, solution planning, code generation, and execution-based validation within a closed loop, thereby serving as a true end-to-end optimization copilot.

Inspired by these insights, this paper proposes a multi-LLM agentic AI architecture that autonomously manages the entire optimization workflow from conception to solution. Unlike prior studies that apply LLMs as subordinate components or isolated formulators, the proposed framework mimics the feedback-driven process of a human research expert. It coordinates specialized agents (e.g., literature-surveying, planning, coding, and scoring agents) to autonomously perceive a natural-language task, formulate the optimization problem, generate numerical simulation code, and, most critically, iteratively execute and verify the solution for correctness, feasibility, and physical reasonability. In this manner, the proposed approach bridges the gap between manual high-level design exploration and the growing need for autonomous, code-centric validation in next-generation dynamic wireless networks.

\section{Agentic AI Framework and Methodology}
\label{sec:framework}

In this section, we describe the architectural framework and operational methodology of the proposed multi-LLM agentic AI system. Different from using an LLM as a stand-alone text generator, our goal is to construct an optimization copilot that is explicitly tailored for wireless network optimization, where decisions should be numerically precise, constraint-intensive, and physically meaningful. In emerging wireless networks (e.g., ISAC, SWIPT, RIS, SAGIN, low-altitude networks, and edge intelligence), the optimization pipeline is often the main bottleneck: New intents should be translated into rigorous models, feasible algorithms, executable code, and verifiable outputs. Therefore, the proposed framework is designed to close this gap by coupling intent understanding with code-centric execution and physics-aware verification in a unified closed loop. 

\subsection{A communication-oriented agentic AI architecture}
\label{subsec:ppar}

Unlike conventional static LLM pipelines, our framework functions as an autonomous cognitive system governed by a recursive Perception--Planning--Action--Reflection cycle that is instantiated as a communication-oriented controller that aligns with the lifecycle of wireless optimization, from problem interpretation to solver execution and feasibility checking.

\textbf{Perception:} Perception serves as the system interface for state awareness, operating continuously rather than as a singular initialization step. From a wireless network perspective, perception must jointly capture semantic intents (e.g., QoS targets, sensing accuracy, energy harvesting requirements, and latency budgets) and engineering states (e.g., channel state information (CSI), statistical channel models, topology, resource budgets, and protocol constraints). At the macro level, it parses the user intent into explicit design constraints such as beamforming objectives, power allocation requirements, and energy guarantees. At the micro level, specialized agents monitor the execution environment and interpret diverse feedback, including retrieval relevance, solver infeasibility reports, numerical instability, and simulation mismatches. This multi-layered perception keeps the workflow grounded in both the communication objective and the evolving runtime state, mitigating open-loop hallucinations. 

\textbf{Planning:} Planning constitutes the core reasoning engine, executing task decomposition recursively across abstraction levels. For wireless network design, planning must translate an intent into a solvable and verifiable optimization pipeline, rather than only producing a formal objective. On a systemic scale, it orchestrates a workflow that breaks a complex task into stages such as channel and system modeling, variable and constraint specification, algorithm selection, baseline construction, and evaluation protocol definition. On an atomic scale, agents decide the sequence of tool invocations (e.g., convex solvers and simulation scripts) and derive intermediate formulations to handle non-convex couplings. This hierarchical planning enables structured exploration in high-dimensional design spaces typical for wireless networks, e.g., mixed discrete and continuous decisions, multi-objective trade-offs, and coupled constraints.

\textbf{Action:} Action represents the capacity to interact with and manipulate the external environment, transforming the LLM from a passive generator into an active executor. For communication tasks, {action} corresponds to tool-integrated operations such as literature retrieval, solver calls, dataset generation, simulation execution, and baseline reproduction. Action is crucial because the correctness of wireless optimization is established through executable validation under physical constraints (e.g., power budgets, harvested-energy thresholds, signal-to-interference-plus-noise ratio (SINR) feasibility, sensing metrics, and numerical stability). Each executed action yields empirical feedback that is ingested by perception to trigger the next cycle of analysis and refinement.

\textbf{Reflection:} Reflection acts as an active critic that identifies hallucinations, logical inconsistencies, and feasibility violations, and iteratively refines strategies and actions. In wireless network optimization, reflection must be physics-aware. That is, it checks whether constraints are satisfied, whether outputs remain physically meaningful (e.g., rank conditions are satisfied or appropriately relaxed, and power and energy constraints are met), and whether the implemented code faithfully follows the planned algorithm. The feedback-driven reflection bridges the probabilistic nature of LLM generation and the rigorous precision required by wireless designs, turning the agent into a resilient, self-correcting orchestrator.

\begin{figure*}[t!]
\centering
 \makebox[\textwidth]{\includegraphics[width=.84\paperwidth]{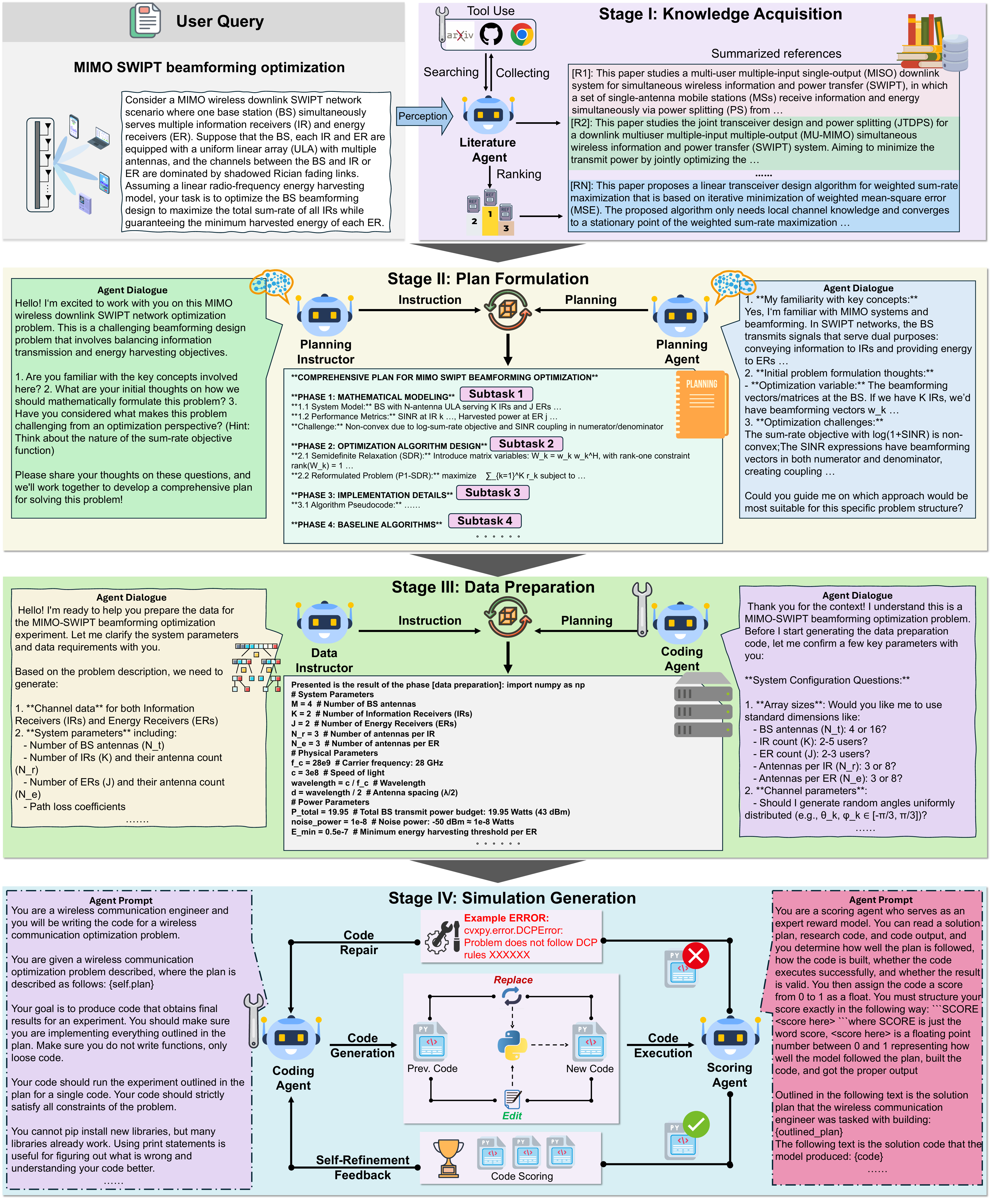}}
\caption{Operational Breakdown of the ComAgent Framework. A detailed example workflow that demonstrates the autonomous execution of a case study on MIMO SWIPT beamforming optimization across four distinct stages.}
\label{stage1+2}
\end{figure*}
\subsection{End-to-end workflow instantiation for network optimization}
\label{subsec:stages}

Based on the above agentic loop, we implement the proposed framework as a four-stage workflow that matches typical wireless network design practices: Knowledge acquisition, plan formulation, data preparation, and simulation generation. The key objective is to produce solver-ready models and execution-verified results under communication constraints, rather than only generating plausible descriptions.

\subsubsection{Knowledge acquisition}
As illustrated in Fig.~\ref{stage1+2}, Stage~I serves as the knowledge grounding module for network optimization. The Literature Agent first parses the natural-language request and identifies the task type and evaluation targets, such as SWIPT beamforming, ISAC co-design, RIS configuration, or SAGIN resource management. It then extracts the essential modeling elements that determine feasibility and performance, including channel assumptions, antenna settings, hardware constraints, and task-specific metrics.

The agent leverages external tools, including the Semantic Scholar Search API and web browsers, to retrieve relevant papers and baselines. Instead of a single search-and-summarize step, we adopt an iterative filtering process. The agent issues multiple queries, examines candidate abstracts and key sections, and refines the candidate list based on alignment with the target scenario and constraints. This design is important for wireless optimization because small modeling differences, such as the SWIPT energy harvesting model or the ISAC sensing metric, can lead to incompatible formulations and misleading comparisons.

After retrieval, the Literature Agent ranks the selected papers by scenario relevance and extracts the most useful components, including standard problem formulations, commonly used constraints, and representative solution methods. The resulting literature knowledge, together with the original user query, is stored in the System Memory and directly supports the next stage.

\subsubsection{Plan formulation}

Stage~II translates the grounded knowledge into a solver-ready optimization plan. A Planning Instructor acts as the coordinator and guides the Planning Agent through iterative refinement. To enhance the reasoning capabilities, agents apply advanced prompting
strategies, including chain-of-thought (CoT), reasoning-action (ReAct), and plan-and-solve (PS) methodologies \cite{zhou}. Specifically, CoT prompting elicits explicit intermediate reasoning steps from LLMs, which can enhance performance for complex reasoning tasks but also increase inference cost and may expose unstable or biased internal rationales. ReAct integrates natural-language reasoning with external tool use and environment interaction in an interleaved loop, enabling the model to iteratively plan, act (e.g., query tools), observe outcomes, and update its reasoning based on feedback. PS first instructs the model to generate a high-level solution plan and then prompts it to carry out this plan step by step, thereby improving global coherence and reducing errors compared with purely incremental, token-by-token solution generation.

The process starts by retrieving the user query and the literature knowledge from the System Memory. The Planning Agent then specifies the system model and design objectives in a form that can be implemented, including optimization variables, constraints, and performance metrics. For our tasks, this step explicitly addresses practical requirements, such as power budgets, SINR guarantees, sensing accuracy targets, and energy harvesting thresholds. The agent also selects a solution route that fits the problem structure and numerical properties.

The Instructor reviews the draft plan to identify missing constraints, ambiguous definitions, or steps that are difficult to implement. The Planning Agent revises the plan based on this feedback, and the loop repeats until the plan is internally consistent and implementation-oriented. As shown in Fig.~\ref{stage1+2}, the final output is a structured execution chain that decomposes the original task into linked subtasks, covering formulation, algorithm design, baseline construction, and experiment settings. The design ensures that downstream stages operate on clear and verifiable targets rather than informal descriptions.

\subsubsection{Data preparation}
Stage~III builds the empirical foundation for the simulation and enforces consistency between the plan and the executable setup. The Data Instructor supervises the Coding Agent to generate or collect data that matches the adopted wireless model, which includes channel realizations, topology generation, parameter settings, and task-specific data when needed.
The stage begins with the Data Instructor retrieving the user query and the finalized plan from the System Memory. The Instructor then requests a data-preparation strategy from the Coding Agent and checks whether the strategy matches the plan, including channel models, antenna configurations, and scenario parameters. After validation, the Coding Agent uses external tools, such as Python scripts for synthetic data generation or API calls for public datasets, to construct a local dataset.

To avoid silent mismatches that frequently break wireless simulations, the Data Instructor performs iterative checks during data generation. It verifies code execution, dimensional consistency, and basic sanity of generated samples, such as path-loss trends and power normalization. If issues occur, the Coding Agent revises the script based on the feedback and repeats the process until the dataset is consistent with the plan. The validated dataset is then stored for use in Stage~IV.

\subsubsection{Simulation generation}

Stage~IV is the execution and validation engine of the framework, as shown in Fig.~\ref{stage1+2}. The Coding Agent converts the solution plan and prepared dataset into executable simulation code in a sandbox workspace. The implementation follows a modular structure so that the system model, solver, baselines, and evaluation metrics can be checked independently. During coding, the agent keeps the implementation aligned with the plan and records modifications to improve traceability.

To emphasize communication relevance, validation is not limited to code execution. The Scoring Agent checks whether the implementation satisfies wireless constraints and whether the produced results are physically meaningful. The evaluation follows two feedback branches.

\begin{itemize} 
\item \textbf{Error handling branch}: If the simulation fails to compile or execute, the Scoring Agent captures the specific error signal (e.g., syntax errors or runtime exceptions) and reports it to the Coding Agent. The Coding Agent then performs self-reflection to diagnose the root cause and rectify the issue in the subsequent iteration.
\item \textbf{Wireless validity branch}: Upon successful compilation, the Scoring Agent evaluates feasibility and result consistency. Typical checks include constraint satisfaction, numerical stability, and whether key metrics behave as expected under parameter changes, for example, rate versus transmit power. The agent employs an LLM-based reward model to compute a quantitative score ranging from 0 to 1. Higher scores indicate greater alignment with the initial task requirements and physical constraints. \end{itemize}

Based on the structured feedback, comprising both the quantitative score and qualitative analysis, the Coding Agent reflects on the current implementation and iteratively refines
the code to maximize the reward. This closed-loop refinement is essential for wireless network optimization tasks, where minor implementation errors or missing constraints can produce results that seem reasonable but violate physical limits. By coupling execution feedback with wireless-specific validity checks, Stage~IV turns the agentic loop into a practical tool for generating reproducible and communication-ready simulations.
\section{Performance evaluation: From autonomous design to robust generalization}
In this section, we present two case studies to demonstrate the effectiveness of the proposed multi-LLM agentic AI system. The Claude-4.5-Sonnet is chosen as the base model for all agents.
\subsection{Case study 1: MIMO SWIPT beamforming optimization}
To demonstrate the autonomous end-to-end orchestration capability of our framework, we task the system with a non-trivial MIMO SWIPT beamforming optimization problem specifically designed by a human expert,  \textcolor{black}{which considers a MIMO wireless downlink SWIPT network that simultaneously serves $K$ information receivers (IR) and $J$ energy receivers (ER) under shadowed Rician fading channels} to maximize the total sum-rate of all IRs while guaranteeing the minimum harvested energy of each ER. The detailed process of employing the proposed multi-LLM agentic AI system to solve the task is as follows.

The process begins with a user query using natural language for the MIMO SWIPT beamforming task as follows:
\begin{mdframed}[backgroundcolor=gray!20, linewidth=0pt]
\textbf{User Query}: \textit{\textcolor{black}{Consider a MIMO wireless downlink SWIPT network scenario where one base station (BS) simultaneously serves multiple information receivers (IR) and energy receivers (ER). Suppose that the BS, IR, and ER are equipped with a uniform linear array (ULA) with multiple antennas, and the channels between the BS and IR or ER are subject to shadowed Rician fading links. {Assuming a linear radio-frequency energy harvesting model, your task is to optimize the BS beamforming design to maximize the total sum-rate of all IRs while guaranteeing the minimum harvested energy of each ER.}}}
\end{mdframed}
\subsubsection{Autonomous end-to-end solution}
To process the user query, the Literature Agent first anchors the task in existing research by retrieving and filtering key references on MIMO SWIPT from external repositories. Leveraging the domain knowledge, the Planning Agent does not merely copy and paste formulas but autonomously collaborates with the Planning Instructor to translate high-level intent into a rigorous mathematical framework. Recognizing the inherent non-convexity of the SWIPT beamforming problem, the Planning Agent strategically decomposes the task into a logical chain: starting from channel modeling, proceeding to algorithm design (specifically transforming non-convex constraints via semidefinite relaxation (SDR) and successive convex approximation (SCA) techniques), and concluding with baseline selection for comparative analysis. A critical insight observed during the experiment is that the Planning Agent demonstrates expert-level reasoning. Recognizing that the coupling between beamforming vectors and energy harvesting constraints renders the problem highly non-convex, the agent does not blindly attempt gradient descent. Instead, it autonomously formulates the problem using SDR and SCA to handle the rank-1 constraints, as shown in Fig. \ref{stage1+2}, which verifies that the agent possesses mathematical reasoning and efficiently utilizes domain knowledge derived from the Knowledge Acquisition stage.

The execution phase is orchestrated through a tight feedback loop between the Coding and Scoring Agents. Guided by the Coding Instructor, the Coding Agent generates a simulation environment. Then, following the solution plan, the Coding Agent generates simulation scripts. Crucially, the Scoring Agent acts as an expert critic, which evaluates the generated solution not only on syntax correctness but also on physical feasibility, and ensures that the semantic constraints, such as the minimum harvested energy threshold and BS transmit power budget, are strictly satisfied. Through this iterative Perception-Planning-Action-Reflection cycle, the system autonomously identifies and rectifies logical errors in the beamforming algorithm, refining the code until it converges to a physically valid and optimized solution.

\subsubsection{Performance verification}
\begin{figure}
    \centering
    \includegraphics[width=1\linewidth]{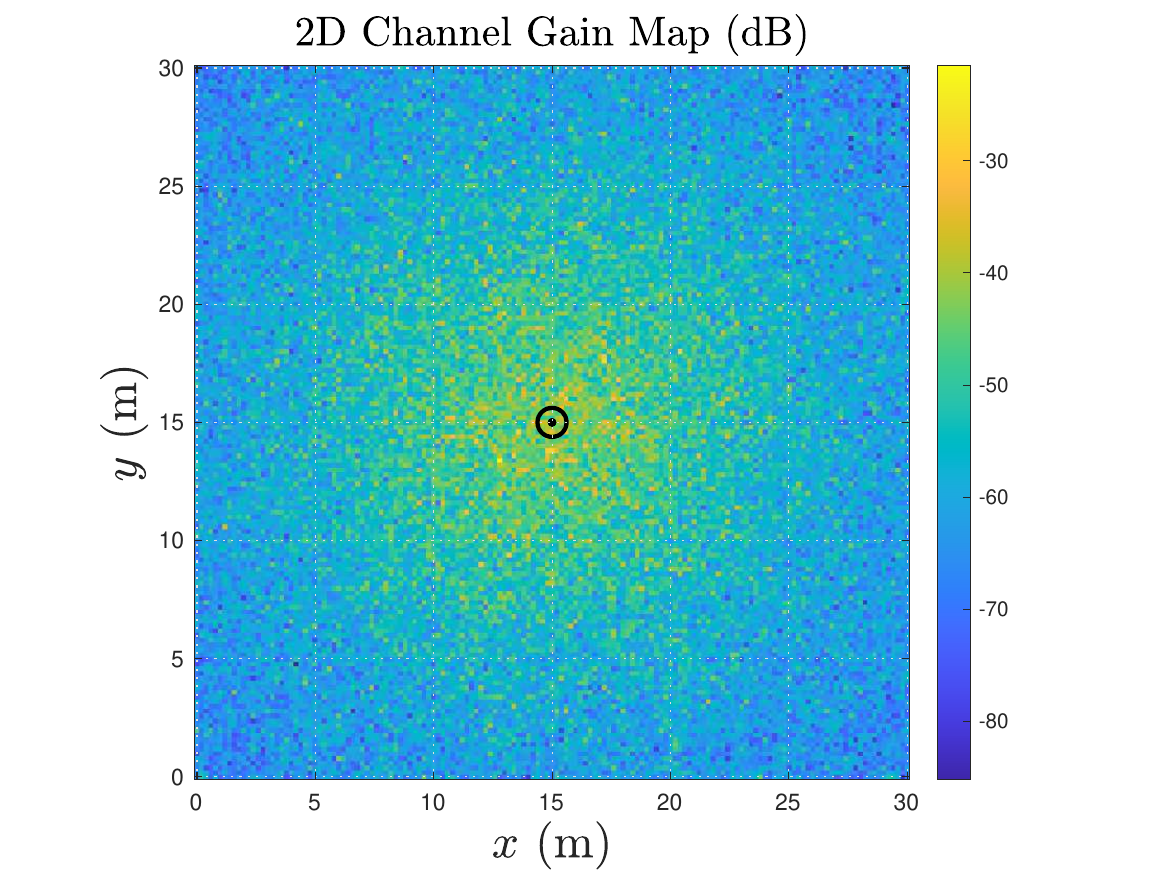}
    \caption{ \textcolor{black}{An example 2D channel gain map with the BS located at the origin.}}
    \label{channel}
\end{figure}
\begin{figure}
    \centering
    \includegraphics[width=1\linewidth]{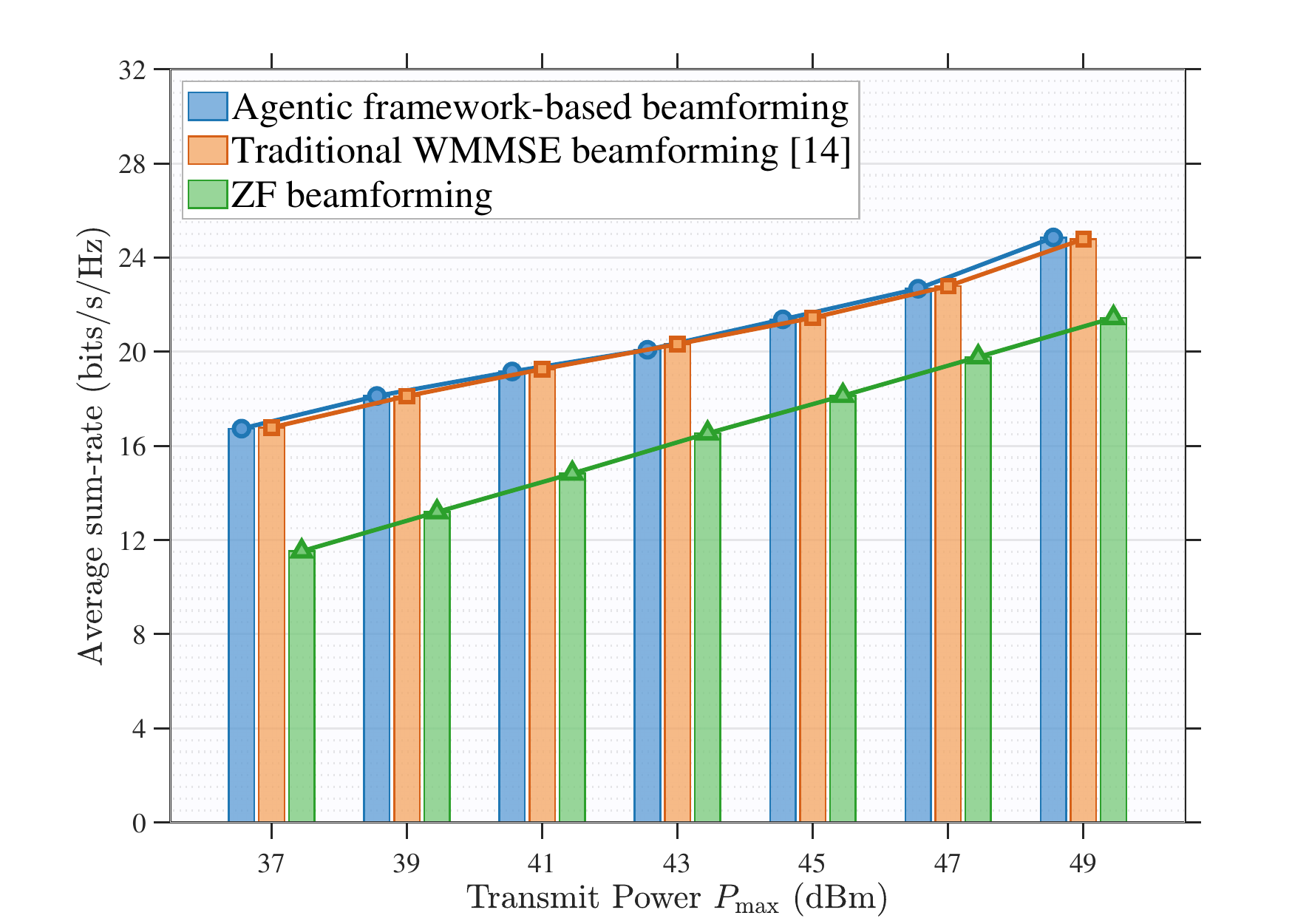}
    \caption{Average sum-rate versus transmit power $P_{\mathrm{max}}$.}
    \label{power}
\end{figure}
\begin{figure}
    \centering
    \includegraphics[width=1\linewidth]{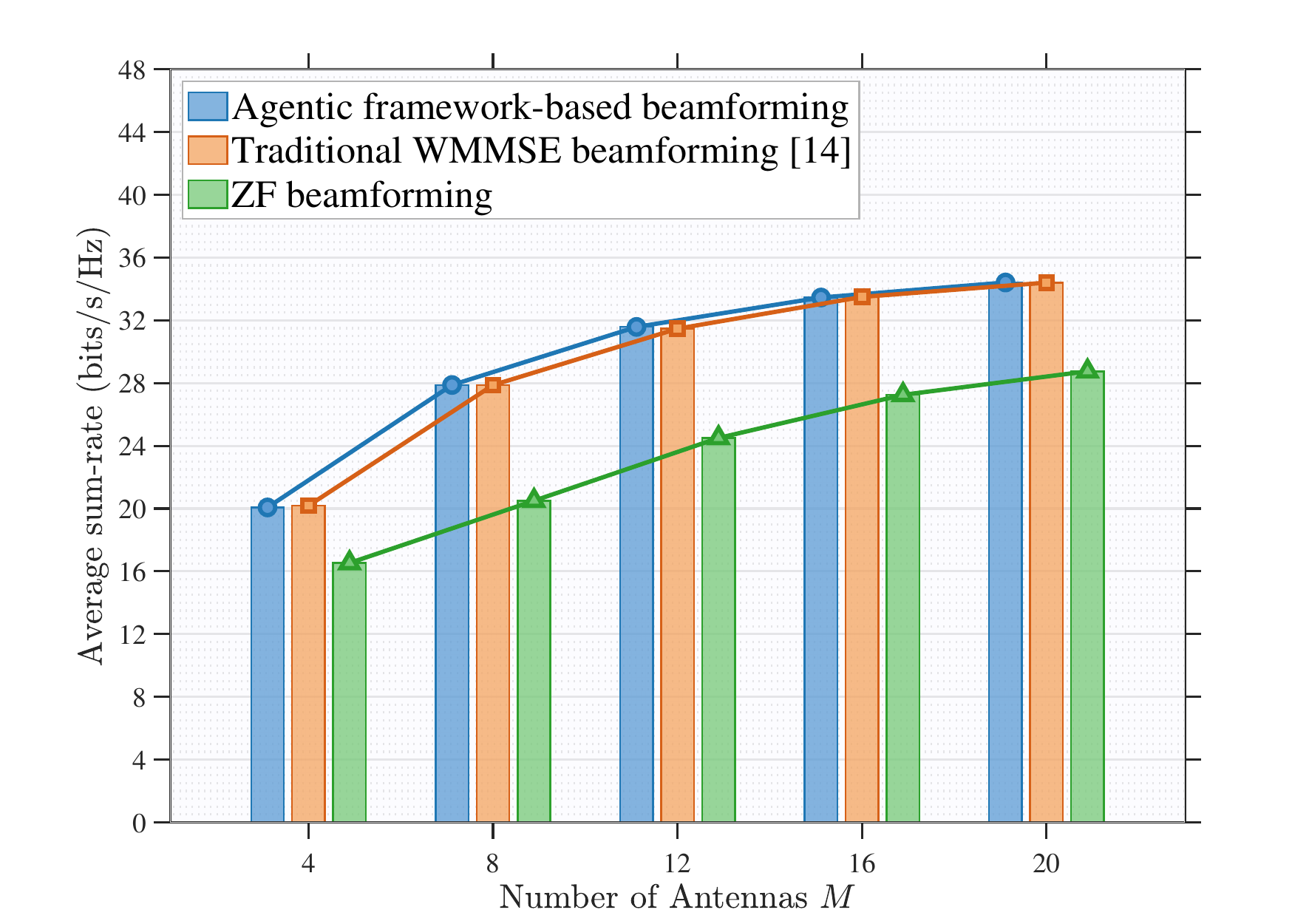}
    \caption{Average sum-rate versus the number of antennas $M$.}
    \label{antenna}
\end{figure}

In this case study, the multi-LLM agentic AI system generates the simulation code, including the system parameter settings, algorithm design, and baseline implementations. Specifically, there are $K=2$ IRs and $J=2$ ERs randomly and uniformly distributed in a two-dimensional (2D) rectangular region of $30$ m $\times$ $30$ m. The system operates at a carrier frequency of $f_c=28$ GHz. The BS is equipped with $M=4$ antennas and is deployed at a height of $h=5$ m. Each IR and ER is equipped with $N_r=N_e=3$ antennas with half-wavelength antenna spacing. \textcolor{black}{The large-scale channel attenuation follows $L(d)=L_0+10\alpha\log_{10}(d/d_0)$ with a reference distance of $d_0=1$ m, a path-loss factor of $L_0=30$ dB, and a path-loss exponent of $\alpha=3.5$. The shadowing is modeled as lognormal with a standard deviation of $\sigma_{\mathrm{sh}}=6$ dB, and the small-scale fading is modeled as Rician with a factor of $K_{\mathrm{rician}}=6$ dB.} The receiver noise power is set to $\sigma^2=-50$ dBm, the maximum transmit power is $P_{\mathrm{max}}=43$ dBm, and the minimum harvested energy requirement is $E_{\mathrm{min}}=0.5\times10^{-7}$ W. To evaluate the optimality of this agent-generated solution, we compare it with two distinct benchmarks: the weighted minimum mean-squared-error (WMMSE) algorithm \cite{wmmse}, which represents a traditional expert-level iterative solution, and the zero-forcing (ZF) beamforming. The simulations are repeated multiple times with different random seed settings and averaged.

\textcolor{black}{Fig. \ref{channel} shows an example 2D channel gain map of shadowed Rician fading channels generated by the agentic AI framework. It is shown that our proposed framework accurately captures the features of the fading channels rather than simplifying them as line-of-sight links.} Fig. \ref{power} illustrates the average achievable sum-rate performance versus the maximum transmit power $P_{\max}$. It is shown that the performance of the agentic framework-based beamforming is comparable to that of the traditional WMMSE algorithm. This alignment indicates that the code generated by the Coding Agent successfully implements the sophisticated iterative logic required to balance signal strength against inter-user interference. Furthermore, the agent-generated solution consistently outperforms the ZF baseline. We further investigate the impact of the spatial degrees of freedom (DoF) on system performance in Fig. \ref{antenna}, which depicts the average sum-rate versus the number of transmit antennas $M$. As $M$ increases, the system performance improves significantly. The results show that the agent-generated algorithm effectively exploits the additional spatial DoF to generate sharper beams. Even in high-dimensional settings, the agentic AI solution maintains its performance advantage over ZF and is comparable to the WMMSE benchmark. These results show that the proposed multi-LLM framework can autonomously explore the design space, select appropriate mathematical tools (e.g., SDR/SCA), and generate robust code that converges to expert-level solutions without human intervention.
\begin{table*}[h]
    \centering
    \caption{Performance comparison on the generic test set}
    \label{tab:performance_comparison}
    \renewcommand{\arraystretch}{1.2} 
    \setlength{\tabcolsep}{10pt} 
    \begin{tabular}{lccc}
        \toprule
        \textbf{Metric} & \textbf{Single LLM} & \textbf{Single LLM + PS} & \textbf{Agentic AI Framework} \\
        \midrule
        Problem Formulation Rate & 0.00\% (0/25) & 56.00\% (14/25) & \textbf{100.00\% (25/25)} \\
        Code Generation Rate & 100.00\% (25/25) & 100.00\% (25/25) & \textbf{100.00\% (25/25)} \\
        Code Execution Rate & 24.00\% (6/25) & 88.00\% (22/25) & \textbf{100.00\% (25/25)} \\
        Solution Solved Rate & 24.00\% (6/25) & 56.00\% (14/25) & \textbf{72.00\% (18/25)} \\
        1st-Try Success Rate & 4.00\% (1/25) & 20.00\% (5/25) & \textbf{32.00\% (8/25)} \\
        Avg. Attempt Times & 2.88 & 2.44 & \textbf{2.12} \\
        \bottomrule
    \end{tabular}
\end{table*}

\subsection{Case study 2: Generalization and robustness across tasks}
To rigorously evaluate the adaptability of the proposed framework across diverse wireless use cases, we established a generic test set comprising 25 specifically human-expert-designed optimization tasks extracted from \cite{liu2}, ranging from classical resource allocation to emerging technologies, such as non-orthogonal multiple access (NOMA) and RIS. We compared the proposed agentic AI framework with two baselines: 1) a single LLM with standard input-output prompting, and 2) a single LLM plus PS strategy, which explicitly prompts the model to generate a plan before coding but lacks iterative environmental feedback. For a fair comparison, the Claude-4.5-Sonnet is chosen as the base LLM, and the model temperature is set to 1.0. The maximum number of attempts per task is set to 3. The performance is quantified using six comprehensive metrics: 1) Problem Formulation Rate, defined as the proportion of tasks in which the system explicitly constructs mathematical models before coding; 2) Code Generation Rate, the ratio of tasks yielding executable scripts; 3) Code Execution Rate, the percentage of generated scripts that run without syntax errors; 4) Solution Solved Rate, the fraction of tasks in which the final output satisfies all constraints and is physically meaningful; 5) 1st-Try Success Rate, indicating zero-shot capability; and 6) Average Attempt Times, representing the mean number of iterations required per task.

The comparative results reveal significant performance disparity of different methods, as summarized in Table \ref{tab:performance_comparison}. The single LLM baseline exhibited a 0\% Problem Formulation Rate, characterizing a leap-to-code failure mode where mathematical modeling is bypassed. This lack of structural reasoning resulted in a catastrophic Code Execution Rate of 24\% and a low Solution Solved Rate of 24\%, as the generated scripts were often syntactically incorrect or logically flawed, such as ignoring physical power budget constraints.

The Single LLM plus PS approach improved performance by enforcing a reasoning step, achieving a 56\% Formulation Rate and raising the Solution Solved Rate to 56\%. However, its static nature limits robustness. Without feedback, it struggled to recover from initial errors, as evidenced by a 1st-Try Success Rate of 20\% and higher Average Attempt Times of 2.44 compared to the agentic AI framework.

In contrast, the agentic AI framework demonstrated superior autonomy and efficiency. It achieved a 100\% Problem Formulation Rate, ensuring that every simulation script was mathematically grounded. Crucially, the closed-loop feedback mechanism allowed the system to self-correct effectively, yielding the highest Solution Solved Rate of 72\%, 1st-Try Success Rate of 32\%, and lowest Average Attempt Times of 2.12. This indicates that the proposed multi-LLM agentic architecture does not merely plan but actively adapts, transforming the LLM from a static generator into a robust, high-precision, and autonomous copilot for complex wireless optimization tasks.

\section{Challenges and Future Directions}
\label{sec:challenges}
Although the proposed multi-LLM agentic AI framework successfully automates specific wireless optimization tasks, transitioning this copilot into a fully autonomous controller for real-world wireless networks introduces practical engineering challenges. In operational network environments, systems must move beyond solving static problems to manage dynamic infrastructures. This section outlines the key architectural evolution required for next-generation agentic AI systems.
\subsection{From task-based execution to continuous operation}
Currently, our framework operates in a one-shot mode in which agents receive a query, execute a plan, and terminate once the solution is generated. Although effective for offline optimization, this episodic workflow must evolve into a continuous service to handle the persistent nature of live wireless networks, where objectives and decision contexts constantly evolve. Future architectures should require agents to maintain a persistent state, continuously monitor network key performance indicators, and dynamically trigger the Planning Agent only when the performance deviates from the targets. This shift transforms the system from a static tool for generating solution plans into a dynamic control loop capable of real-time policy maintenance and intent reconciliation, ensuring that the system can detect deviations without deriving the entire reasoning chain from scratch.

\subsection{Balancing reasoning capabilities with real-time latency}
The current framework relies on extensive interactions and feedback loops between agents, such as the debate between the Planning and Instructor agents, which introduces computational latency that is incompatible with millisecond-level wireless control tasks. A hierarchical control architecture is necessary to bridge the gap between the slow inference speed of large models and the strict timing requirements of the physical layer. Future multi-LLM agentic AI systems should function as a high-level control plane for intent understanding and strategic planning while delegating time-critical execution to lightweight, specialized small models or conventional controllers. This architecture ensures that the comprehensive reasoning of agents enhances network management without becoming a bottleneck for signal processing latency, allowing the system to maintain stability even when the control alternates between components with different response times.

\subsection{Coordination scalability in distributed architectures}
As the framework scales to support large and complex networks, the coordination cost of exchanging lengthy prompts and feedback between agents becomes a critical bottleneck. Relying on a centralized orchestrator for all communications creates a single-point-of-failure bottleneck and high signaling overhead, whereas fully decentralized approaches risk inconsistency. To address the challenge, future systems must adopt adaptive agent topologies in which agents form dynamic subgroups based on task requirements rather than maintaining global connectivity. Furthermore, optimizing communication protocols to exchange structured data or compact tokens instead of full natural language text will be essential to ensure efficient collaboration as network complexity increases, and prevent coordination overhead from negating the benefits of multi-agent intelligence.

\subsection{Long-term memory and lifelong adaptation}
In the current case studies, the proposed framework relies on the Literature Agent to retrieve knowledge anew for each task, lacking a mechanism to retain long-term experience across sessions. Since wireless environments are nonstationary and dynamic, this leads to inefficient duplication of efforts for recurring optimization problems. Future research should be able to integrate long-term experiential memory, distinct from temporary system memory, to archive valid solution templates and failure cases. By retrieving validated strategies directly from this historical archive, the Planning Agent can bypass time-consuming design loops for familiar scenarios, enabling the system to progressively adapt and improve its robustness over its operational lifecycle, rather than remaining perpetually reactive.

\section{Conclusions}
\label{sec:conclusion}
We proposed ComAgent, a multi-LLM based agentic AI framework designed to autonomously navigate the complex design space of emerging wireless networks, addressing the complexity limitations of traditional manual optimization. By orchestrating specialized agents like Literature, Planning, Coding, and Scoring Agents through a proposed recursive Perception-Planning-Action-Reflection cognitive cycle, the framework effectively mitigated the hallucination and reasoning deficits inherent in monolithic LLMs. Extensive case studies, including MIMO SWIPT beamforming and a generic optimization test set, validated that the system not only automated the end-to-end workflow, from mathematical modeling to simulation implementation, but also achieved a solution quality comparable to expert-designed baselines. Consequently, this paper established a foundational multi-LLM agentic AI architecture for intent-driven network management, paving the way for future self-evolving and zero-touch 6G network systems.

\end{document}